\documentclass{article}

\usepackage[T1]{fontenc}    % use 8-bit T1 fonts
\usepackage{url}            % simple URL typesetting
\usepackage{amsfonts}       % blackboard math symbols
\usepackage{nicefrac}       % compact symbols for 1/2, etc.
\usepackage{microtype}      % microtypography
\usepackage{enumitem}
\usepackage{lmodern}
\usepackage{times}
\usepackage{titlesec}
\usepackage{xcolor, color, soul}
\sethlcolor{yellow}

\usepackage{microtype}
\usepackage{graphicx}
\usepackage{subfigure}
\usepackage{booktabs} % for professional tables
\usepackage{hyperref}
\usepackage{url}

\usepackage[accepted]{icml2023}

\usepackage{amsmath}
\usepackage{amssymb}
\usepackage{mathtools}
\usepackage{amsthm}

\usepackage[capitalize,noabbrev]{cleveref}

\theoremstyle{plain}
\newtheorem{theorem}{Theorem}[section]
\newtheorem{proposition}[theorem]{Proposition}

\theoremstyle{definition}

\theoremstyle{remark}
\newtheorem{remark}[theorem]{Remark}

\usepackage[textsize=tiny]{todonotes}

\icmltitlerunning{EBLIME: Enhanced Bayesian Local Interpretable Model-agnostic Explanations}

\begin{document}

\twocolumn[
\icmltitle{EBLIME: Enhanced Bayesian Local Interpretable Model-agnostic Explanations}



\icmlsetsymbol{equal}{*}

\begin{icmlauthorlist}
\icmlauthor{Yuhao Zhong}{y}
\icmlauthor{Anirban Bhattacharya}{yy}
\icmlauthor{Satish Bukkapatnam}{equal,y}
\end{icmlauthorlist}

\icmlaffiliation{y}{Department of Industrial \& Systems Engineering, Texas A\&M University, College Station, TX, USA}
\icmlaffiliation{yy}{Department of Statistics, Texas A\&M University, College Station, TX, USA}

\icmlcorrespondingauthor{Satish Bukkapatnam}{satish@tamu.edu}

\icmlkeywords{Bayesian hierarchical model, Explainable AI (XAI), object localization, ridge parameter, uncertainty quantification}

\vskip 0.3in
]



\printAffiliationsAndNotice{}  % leave blank if no need to mention equal contribution
% \printAffiliationsAndNotice{} % otherwise use the standard text.

\begin{abstract}
We propose EBLIME to explain black-box machine learning models and obtain the distribution of feature importance using Bayesian ridge regression models. We provide mathematical expressions of the Bayesian framework and theoretical outcomes including the significance of ridge parameter. Case studies were conducted on benchmark datasets and a real-world industrial application of locating internal defects in manufactured products. Compared to the state-of-the-art methods, EBLIME yields more intuitive and accurate results, with better uncertainty quantification in terms of deriving the posterior distribution, credible intervals, and rankings of the feature importance. 
\end{abstract}

\section{Introduction}
With an estimated annual growth rate of 51.1\%, the market of deep learning is set to reach \$415B by 2030 \cite{acumen2022}. This rapid growth is due to deep learning models’ ability to make accurate predictions, especially in domains where the underlying knowledge (e.g., process physics) is unavailable or unsettled. Such prediction accuracies are owing to the model's ability to learn the key physical relationships and underlying domain knowledge. However, much of the knowledge learned by these models is hard to discern and remains undiscovered. This has affected the trustworthiness of these models, hampering their broader applicability. Therefore, it is necessary to develop methods to explain and rationalize the black-box model predictions \cite{zhong2022}.

To formalize, let us consider a black-box model $f(\cdot)$ that takes an input $x \in \mathbb{R}^{p}$, with $p$ features. ``Explanation'' in the present context involves providing insights on how the $p$ features in $x$ affect the prediction $f(x)$. To this end, multiple post-hoc methods have been developed to explain black-box models after they have been built. A main branch of the methods attempts to train an interpretable surrogate model (e.g., linear regression) to approximate the black-box model at a local region in the feature space. In this way, the surrogate model (e.g., coefficients of a linear regression model, denoted as $\beta \in \mathbb{R}^{p}$) can be used to represent the local knowledge (e.g., feature importance) of the black-box model. Here, ``local'' is defined by a subset (also called a perturbed dataset) of the neighborhood of the given input $x$.

Local Interpretable Model-agnostic Explanation (LIME) \cite{lime} is one of such techniques that has been widely applied in domains including law, business, healthcare, and manufacturing \cite{gramegna2021,zafar2019dlime,zhong2022}. However, it inherently has a large computational complexity in generating all the possible perturbed datasets. To reduce the computational cost, LIME is conventionally implemented with random sampling and approximation (elaborated in \cref{prelim}). Such treatment induces uncertainty and inconsistent results. The generated explanation regarding the same $x$ and $f$ can be drastically different depending on the random initializations. 

Researchers have attempted to reduce the uncertainty in LIME by making the process deterministic (e.g., DLIME \cite{zafar2019dlime} and SLIME \cite{zhou2021}) or by simply combining and averaging over multiple explanations \cite{zhang2019}. These solutions, however, yield merely a point estimate of the feature importance and may still generate inconsistent results. To naturally take the uncertainty into account, a few studies improved LIME from a Bayesian perspective. In \cite{visani2022}, a Bayesian metric was created to assess the consistency among results from repetitive runs with different randomizations. Saini and Prasad \yrcite{saini2022} applied active learning to generate the perturbed instances where the acquisition function includes the variance inference obtained from Gaussian process modeling. Guo et~al. \yrcite{guo2018} employed Bayesian non-parametric regression mixture model to globally approximate the black-box function. Zhao et~al. \yrcite{zhao2021} used Bayesian linear regressors as the local surrogate models and compared results under various parameter settings of the Gaussian priors. Nevertheless, the focus of these studies was not on quantifying the uncertainty in the feature importance.

Recently, Slack et~al. \yrcite{bayeslime} proposed BayesLIME which is the first method to quantify the uncertainty in LIME. But its performance can be limited by the strong assumption that $\beta$ and the local surrogate model error $\epsilon$ share the same covariance. In particular, this may discount or exaggerate the uncertainty in $\beta$ and produce improper explanation results. 

\begin{figure}[hbt]
\begin{center}
\centerline{\includegraphics[width=\columnwidth]{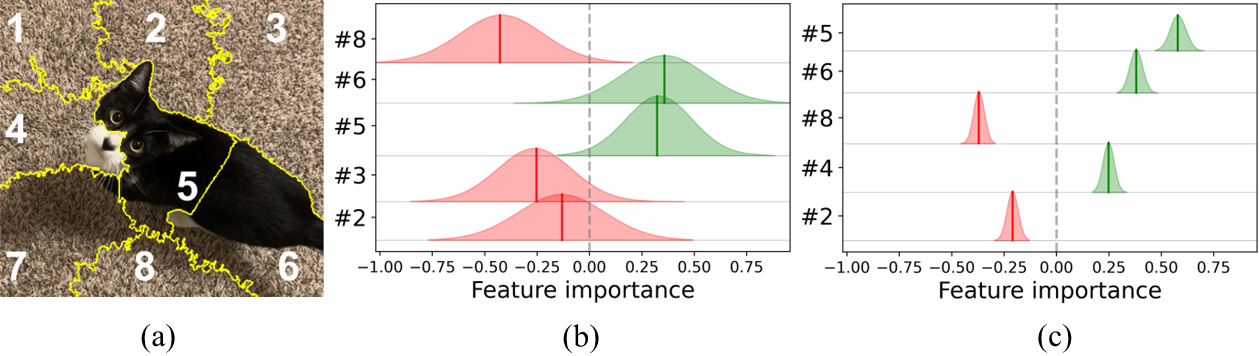}}
\caption{An example of EBLIME explanations. The ``cat'' image in (a) contains 8 non-overlapping segments referred to as \emph{superpixels} (i.e., $p=8$). In (b) and (c), the vertical lines and shaded regions are highlighted in red or green if the mean of the feature importance is negative or positive, respectively. The shaded regions visualize the estimated uncertainty of the superpixel importance. In (b), the explanation is computed using a small perturbed dataset and thus has wider uncertainty intervals \cite{bayeslime}. Due to the overlapping uncertainty, simply relying on the mean may not correctly identify which superpixel is more important. (c) shows the explanation with a larger perturbed dataset. Based on the tighter uncertainty intervals, it is clear that superpixel \#5, which contains the cat head, is the most positively important for the prediction ``cat''.}
\label{sonic}
\end{center}
% \vskip -0.1in
\end{figure}

In this paper, we present EBLIME which improves from BayesLIME. Both BayesLIME and EBLIME aim to address the aforementioned uncertainty while preserving a reasonable computational cost. This is achieved by constructing a Bayesian regression model to approximate $f$ at a local neighborhood around $x$ in the feature space. Coefficients $\beta \in \mathbb{R}^{p}$ of the regression model then indicate the importance of the corresponding $p$ features in $x$. Both approaches inherently elucidate the distributions of $\beta$ and thus helps us determine the true important feature more confidently. Addtionally, in contrast with BayesLIME, EBLIME employs a random variable $\lambda$ to scale the covariance of $\beta$. Thus, it addresses the limitations in BayesLIME. An example of EBLIME result is illustrated in \cref{sonic}.

% Additionally, few studies have discussed about ranking feature importance based on its distributions instead of a point estimate (e.g., mean value). In this regard, we propose an alternative feature ranking criterion for Bayesian explanation methods.

The main contribution of this paper is that we modeled the ridge parameter $\lambda$ as a random variable in the Bayesian regression model \cite{bishop2006}. $\lambda$ scales the covariance of the feature importance $\beta$ so that it differs from the covariance of the model error $\epsilon$. This addressed the strong assumption in existing Bayesian explanation methods that $\textrm{Cov}(\beta)$ and $\textrm{Cov}(\epsilon)$ are of the same scale. In other words, even when the uncertainty of $\epsilon$ is small, the uncertainty of $\beta$ can be large because of variance inflation \cite{james2013}. On the other hand, when the uncertainty of $\epsilon$ is large, the uncertainty of $\beta$ can be small. This is because the generated perturbed datasets may share a similar pattern while containing different data points. 
    % \item We adopted a criterion for sorting the features based on the distribution of their importance instead of a point estimate. The criterion measures the stability and magnitude of the feature importance and does not require any parameter tuning. Here, stability is defined as the probability that the feature importance remains within a fixed value range.
% \end{itemize}

The remainder of the paper is organized as follows: we provide an overview of LIME and its drawback in \cref{prelim}. \cref{methodology} presents the Bayesian framework of EBLIME, including the priors and posterior inference. This is followed by two case studies in \cref{case study}. Conclusions are given in \cref{conclusion}.

\section{Preliminaries} \label{prelim}
As mentioned earlier, LIME is among the most popular techniques for explaining how a trained model makes a prediction on a given input. The explanations usually take the form of interpretable representations, such as feature importance for tabular data, superpixel importance for image data, and word importance for textual data. In this paper, we focus on image data. 

Let us consider a complex classification model $f : x \mapsto [0,1]$, i.e., $f$ maps $x$ to a probability. For convenient illustration, we denote the input $x \in \mathbb{R}^p$ to be an image consisting of $p$ superpixels. The LIME procedure can be generally divided into two steps: perturbation and model fitting.

\textbf{Perturbation}\hspace{3pt} In order to generate interpretable representations of the $p$ superpixels (i.e., features), a local linear surrogate model is built in an alternative feature space. Specifically, LIME creates two new datasets: a perturbed dataset $Z\in {\{0,1\}}^{N\times p}$ in the alternative feature space and its counterpart $Z'\in \mathbb{R}^{N\times p}$ in the original feature space. Both $Z$ and $Z'$ contain $N$ instances with $p$ features. 

In the case of image data, $Z$ and $Z'$ are generated based on the current input $x$. The counterpart of $x$ in the alternative feature space is denoted as $Z_0$ which is a $p$-dimensional vector with all elements equal to 1. Then, $Z_{ij}$ is randomly set to be 0 or 1 according to Bernoulli(0.5), for $i=1,...,N$ and $j=1,...,p$. For each instance $Z'_i$ in the original image space, the pixel intensities within its $j^\textrm{th}$ superpixel are all set to a fixed value (e.g., 0) if $Z_{ij}=0$, or are kept the same as in $x$ if $Z_{ij}=1$. Once the datasets are created, $Z'$ is fed into $f$ to get the black-box predictions $Y\in {[0,1]}^{N}$.

\textbf{Model fitting}\hspace{3pt} In order for the linear model to focus on capturing the local behavior of $f$ around $x$, each instance $Z_i$ is weighted by $\pi_x(Z_i)$. The weight $\pi_x(Z_i)$ is calculated by $\exp{(-D^2(Z_i,Z_0)/\theta^2)}$, which is a kernel with width $\theta$ and a distance measure $D$ (e.g., Euclidean distance between $Z_i$ and the alternative representation of the original input $Z_0$).

Overall, the linear model is fitted by solving a weighted ridge regression, where a constant ridge parameter $\lambda>0$ is specified by the user to penalize the model complexity. The closed-form solution for the model coefficient $\hat{\beta}\in \mathbb{R}^p$ can be written as 

\begin{equation} \label{point_beta}
    \hat{\beta} = V_\lambda (Z^T\textrm{diag}(\Pi_x(Z))Y)
\end{equation} 

wherein $V_\lambda = (Z^T\textrm{diag}(\Pi_x(Z))Z + \lambda\mathbb{I}_p)^{-1}$ and $\textrm{diag}(\Pi_x(Z))$ is a diagonal matrix with elements $\pi_x(Z_i), i = 1,...,N$. $\mathbb{I}_p$ is a $p\times p$ identity matrix. Note that the presence of $\lambda$ also ensures the invertibility of $V_\lambda$. Finally, the magnitude and sign of the coefficients in $\hat{\beta}$ indicate the positive/negative importance of the corresponding input features.

It can be observed that there are a total number of $2^p$ possible perturbed instances, making the problem extremely computationally expensive when $p$ is large. The conventional solution is to use a random subset of all possible perturbations, i.e., $N\ll 2^p$, to approximate the ground truth $\hat{\beta}$. This is rather adhoc and induces uncertainty in the resulting explanations since $\hat{\beta}$ in \cref{point_beta} can vary drastically with different $Z$. To preserve a reasonable computation cost while managing the induced uncertainty, we propose a Bayesian hierarchical method in the next section. 

\section{Enhanced Bayesian local interpretable
model-agnostic explanations}\label{methodology}
\subsection{Prior assumptions} 
Given a perturbed dataset $Z$ and the corresponding black-box prediction $Y$, our proposed explanation model has the following hierarchical form
\begin{subequations}
\begin{gather}
    Y|Z,\beta,\epsilon \sim Z\beta+\epsilon \label{eq2sub1}\\
    \epsilon|\sigma^2 \sim \mathcal{N}(0,\textrm{diag}^{-1}(\Pi_x(Z))\sigma^2) \label{eq2sub2}\\
    \beta|\sigma^2, \lambda \sim \mathcal{N}(0,\lambda^{-1}\sigma^2\mathbb{I}_p) \label{eq2sub3}\\
    \lambda^{-1/2} \sim \textit{\textrm{half-Cauchy}}(0,1) \label{eq2sub4}\\
    \sigma^2 \sim \textit{\textrm{Inverse-Gamma}}(a,b) \label{eq2sub5} \\
    P(\lambda,\sigma^2)=P(\lambda)P(\sigma^2) \nonumber
\end{gather}
\end{subequations}
wherein $\epsilon$ is the error of using the Bayesian ridge regression model to approximate the black-box model. Here, each element of $\epsilon$ is modeled as a zero-mean Gaussian noise with variance inversely related to the weight. In other words, the error is more uncertain for perturbed instances that are farther from the original input $x$. 

It is important to note that in BayesLIME \cite{bayeslime}, both $\epsilon$ and $\beta$ are conditioned on $\sigma^2$ which captures the underlying uncertainty. Their intuition is that the smaller the uncertainty of the error is, the more confident we are supposed to be about the resulting feature importance. Therefore, BayesLIME simply assumed $\beta$ and $\epsilon$ to have the same covariance (the effect of the constant weight matrix in \cref{eq2sub2} can be ignored). This, however, can discount or exaggerate the uncertainty of $\beta$. To address this problem, we introduce a random variable $\lambda>0$ which effectively scales the covariance of $\beta$ and works as a ridge parameter. 

For $\sigma^2$, similar to BayesLIME, we choose a weakly informative inverse-gamma conjugate prior with parameters $a$ and $b$, i.e., $P(\sigma^2)\propto(\sigma^2)^{-a-1}\exp(-b/\sigma^2)$. However, we use a larger value (e.g., 1) for $a$ and $b$, instead of $10^{-6}$ in BayesLIME. This results in heavier-tailed prior and posterior densities that assign less mass near $\sigma^2=0$, allowing larger values of $\sigma^2$ to be plausible \cite{sigmaprior}. Johndrow et~al. \yrcite{johndrow2020} also recommended such heavier-tailed priors on $\sigma^2$, albeit in a sparse regression context. 

For the ridge parameter $\lambda$, we use a default half-Cauchy prior on the scale $\lambda^{-1/2}$. By change of variables, this gives us $P(\lambda)\propto\lambda^{-1/2}(1+\lambda)^{-1} \mathbf{1}_{(0,\infty)}(\lambda)$ . As suggested in \cite{sigmaprior, polson2012}, such a prior %leads to a more proper posterior compared to the 
has better frequentist operating characteristics compared to the commonly adopted conjugate inverse-gamma prior. It is also heavier-tailed, allowing the variability of $\lambda$ to be either very small or large. 

\subsection{Inference} 
Our goal is to obtain the marginal posterior of $\beta$, so that we can analyze the uncertainty of the feature importance given a specific perturbed dataset $Z$ and $Y$.

\begin{proposition} \label{prop0}
Using Bayesian conjugacy adapted to the weighted ridge regression setup, we can write
\begin{gather}
    \beta|\sigma^2, \lambda, Y \sim \mathcal{N}(\hat{\beta}, V_\lambda \sigma^2) \label{beta_cond_pos}\\
    \sigma^2|\lambda,Y \sim \textit{\textrm{Inverse-Gamma}}(a+N/2,Q_{\lambda}/2) \label{sigma_cond_pos}
\end{gather}
wherein $Q_{\lambda} = Y^T\big(\normalfont{\textrm{diag}}^{-1}(\Pi_x(Z))+\lambda^{-1}ZZ^T\big)^{-1}Y + 2b$. 
\end{proposition}

The proof is included in \cref{proofs}. Based on the conditional posterior of $\beta$, we can estimate the marginal posterior mean $E(\beta|Y)$ and covariance $\textrm{Cov}(\beta|Y)$ using \cref{alg1}. The marginal posterior provides insights into the distribution of the feature importance $\beta$, instead of merely a point estimation provided by LIME. To implement \cref{alg1}, we further derive \cref{prop2}.

\begin{algorithm}[htb]
   \caption{Estimate $E(\beta|Y)$ and $\textrm{Cov}(\beta|Y)$}
   \label{alg1}
\begin{algorithmic}[1]
   \STATE {\bfseries Input:} $P(\lambda|Y), P(\sigma^2|\lambda,Y), P(\beta|\sigma^2,\lambda,Y)$, number of posterior samples $s$
   \FOR{$i=1$ {\bfseries to} $s$}
   \STATE Sample $\lambda_i$ from $P(\lambda|Y)$ 
   \STATE Sample $\sigma^2_i$ from  $P(\sigma^2|\lambda_i,Y)$
   \STATE Sample $\beta_{(i)}$ from  $P(\beta|\sigma^2_i,\lambda_i,Y)$
   \ENDFOR
   \STATE {\bfseries return} $E(\beta|Y)$ and $\textrm{Cov}(\beta|Y)$ based on $\{\beta_{(1)},...,\beta_{(s)}\}$
\end{algorithmic}
\end{algorithm}

% \begin{proposition} \label{prop1}
% Given a $\lambda$, 
% \begin{flalign} \label{sigma_cond_pos}
% \sigma^2|\lambda,Y \sim \textit{\textrm{Inverse-Gamma}}(a+N/2,Q_{\lambda}/2), 
% \end{flalign}
% wherein $Q_{\lambda} = Y^T(M_{\lambda})^{-1}Y + 2b$.
% \end{proposition}
% \begin{proof}
% $P(\sigma^2|\lambda,Y) \propto P(Y|\sigma^2,\lambda) P(\sigma^2) \propto (\sigma^2)^{-(a+N/2)-1}\exp[-(Y^TM_{\lambda}^{-1}Y+2b)/(2\sigma^2)]$.
% \end{proof}

\begin{proposition} \label{prop2}
The posterior density of $\lambda$ can be written as
\begin{flalign}  
    \begin{aligned} \label{lam_pos}
    P(\lambda|Y)
    & \propto |M_{\lambda}|^{-1/2}(Q_{\lambda})^{-(a+N/2)}P(\lambda).
    \end{aligned} 
\end{flalign}
wherein $M_\lambda=\normalfont{\textrm{diag}}^{-1}(\Pi_x(Z))+\lambda^{-1}ZZ^T$. The posterior mean of $\lambda$ can be approximated by discretizing $P(\lambda|Y)$ over $\{\lambda_1,...,\lambda_L\}$, that is
\begin{equation} \label{lam_pos_mean}
    {E(\lambda|Y)} \approx \sum^{L}_{l=1}\lambda_l|M_{\lambda_l}|^{-1/2}(Q_{\lambda_l})^{-(a+N/2)}P'(\lambda_l)
\end{equation} 
 % $|\cdot|$ denotes the matrix determinant
wherein $L$ is the number of discretized $\lambda$ values and $P'(\lambda)$ denotes the normalized $P(\lambda)$ satisfying $\sum^{L}_{l=1}P'(\lambda_l)=1$.
\end{proposition}

\begin{remark}\label{r1} An important observation is that in \cref{lam_pos_mean}, $M_\lambda$ and $Q_\lambda$ are functions of $Z$. This means $E(\lambda|Y)$ varies with different perturbed datasets $Z$, not to mention different inputs $x$. Therefore, existing methods that simply set $\lambda$ to be a constant (i.e., $\beta$ and $\epsilon$ sharing a covariance of the same scale) may produce problematic results.
\end{remark}

\begin{remark}\label{r2}  Sampling $\lambda$'s from $P(\lambda|Y)$ can be difficult when $N$ is large because the magnitude of $Q_{\lambda}^{-(a+N/2)}$ can be numerically troublesome. Therefore, Gumbel trick \cite{maddison2016} is employed (see \cref{alg2}).

\begin{remark}\label{r3} Guidance on setting the upper bound $r$ when creating the discretization set $\{\lambda_1,...,\lambda_L\}$: based on the shape of $P(\lambda)$ over $\lambda\in(0,\infty)$, $r$ can be determined such that $P(r)$ is less than a threshold close to 0. 
\end{remark}

\begin{algorithm}[htb]
   \caption{Sample $\lambda$'s from $P(\lambda|Y)$ using Gumbel trick}
   \label{alg2}
\begin{algorithmic}[1]
   \STATE {\bfseries Input:} discretization set size $L$, posterior sample size $s$
   \STATE Initialize $S = \{\}$ as the set of sampled $\lambda$.
   \STATE Suppose $P(\lambda|Y)$ is discretized over $\{\lambda_1,...,\lambda_L\}$.
   \FOR{$i=1$ {\bfseries to} $s$}
   \FOR{$l=1$ {\bfseries to} $L$}
   \STATE $g_l = \log{P'(\lambda_l)}-(1/2)\log{|M_{\lambda_l}|}-(a+N/2)\log{Q_{\lambda_l}}$
    \STATE Draw $\mu_l \overset{\mathrm{i.i.d}}{\sim} \textit{\textrm{Exponential}}(1)$
    \STATE $U_l = -\log{\mu_l}$
   \ENDFOR
   \STATE $l^* = \arg\max_{l\in\{1,...,L\}}(g_l+U_l)$
   \STATE $S.\mathrm{append}(\lambda_l^*)$
   \ENDFOR
   \STATE {\bfseries return} $S$
\end{algorithmic}
\end{algorithm}
\end{remark}

\section{Case studies and results}\label{case study}
Besides the theoretical derivations, we also provide implementation results using the propositions. In the first case study, we compare the uncertainty quantification performance of EBLIME and BayesLIME on the benchmark ImageNet dataset \cite{deng2009}. In the second case, we apply EBLIME to solve a real-world manufacturing problem---to locate the internal defects in products using their ultrasonic scans. Lastly, based on the two case studies, we again emphasize the significance of modeling the ridge parameter $\lambda$. 
\subsection{ImageNet dataset}
In this case study, we employ a pretrained VGG16 convolutional neuron network (CNN) classifier \cite{simonyan2014} as the black-box model. It has been trained to classify over 14 million images into one thousand classes.
\subsubsection{Posterior mean and variance of feature importance}
\begin{figure}[hbt]
\begin{center}
\centerline{\includegraphics[width=\columnwidth]{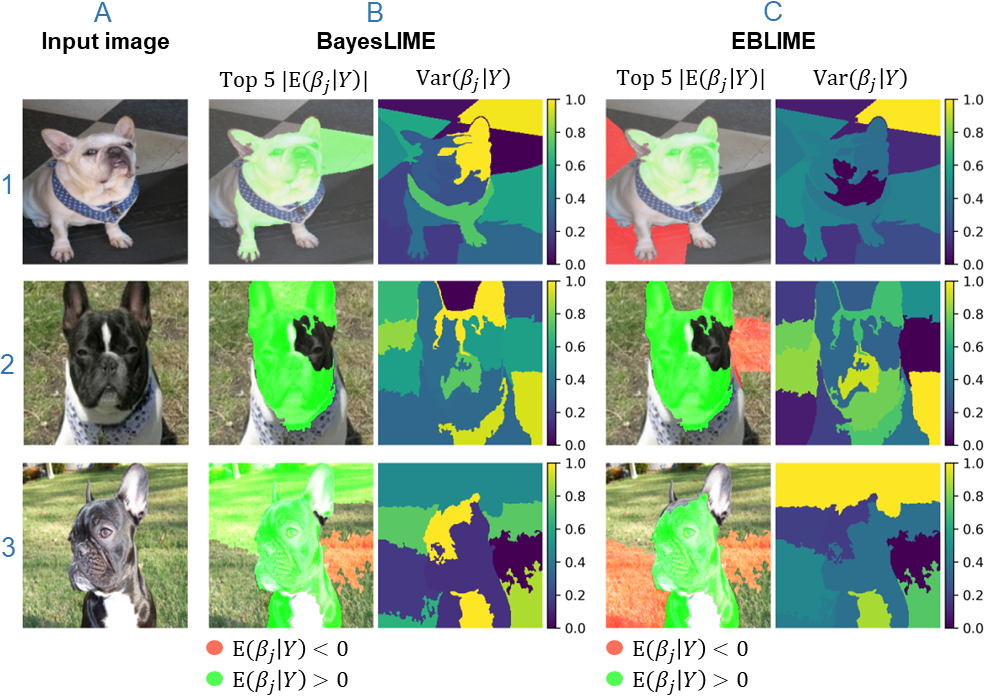}}
\caption{The input images (Column A) and explanation results from BayesLIME (Column B) and EBLIME (Column C). Each explanation result contains two graphs: the left one highlights the 5 superpixels that correspond to the top 5 largest $|E(\beta_j|Y)|$, for $j = 1,...,p$ ; the right one shows the uncertainty of the superpixel importance, i.e., $\textrm{Var}(\beta_j|Y)$.}
\label{pos_m_var}
\end{center}
% \vskip -0.1in
\end{figure}

 \cref{pos_m_var} exemplifies the BayesLIME (Column B) and EBLIME (Column C) explanation results of three images from the class “french bulldog”. The CNN model correctly predicts these images (shown in Column A) as ``french bulldog'' with probabilities 1, 0.98, and 0.99, respectively. This implies that the CNN model most likely has learned the ``french bulldog'' patterns in these images. While explaining these images, the perturbation dataset consisted of 200 instances (i.e., $N=200$) and was kept the same for both BayesLIME and EBLIME. Additionally, we used the default setting of BayesLIME. For EBLIME, $\lambda$ was discretized by uniformly sampling 20,000 values from $(0,1]$, namely $r$ = 1 and $L$ = 20,000. The number of posterior samples $s=$2500. 

To delineate the importance of the superpixels, we highlight the top 5 superpixels that correspond to the five largest components by absolute value in $E(\beta|Y)$. Furthermore, a superpixel is highlighted in green if its corresponding coefficient is positive, and red otherwise. In other words, the presence of a green/red superpixel tends to increase/decrease the probability of the image being predicted as ``french bulldog''. For example, according to EBLIME, the presence of the dog head superpixel in the first image (Row 1, Column C) tends to increase the probability of the image being predicted as ``french bulldog'', while the presence of the background superpixel tends to reduce it. We also show the variance of the importance of each superpixel, i.e., $\textrm{Var}(\beta_j|Y)$ for $j = 1,...,p$. For convenience, the variances were min-max scaled within the result of BayesLIME and EBLIME, respectively. 

For all three images, the top 5 superpixels identified by both EBLIME and BayesLIME correctly included the dog body as positively important. However, BayesLIME assigned a larger variance to some of these dog superpixels. For example, BayesLIME results summarized in Column B show the face of the dogs in Rows 1 and 3, as well as the right ear of the dog in Row 2 in yellow. This means BayesLIME is not as confident about the correct explanation as EBLIME does. Moreover, BayesLIME tended to incorrectly consider the background superpixel around the dog as positively important with a small variance (i.e., higher confidence) for all three images. This is not the case for EBLIME (see \cref{pos_m_var}, Column C) as it did not include these background superpixels in the top 5. Instead, EBLIME correctly outlined other background superpixels as negatively important with a small variance in all three images.

The improper outcomes from BayesLIME were possibly caused by the existence of multicollinearity \cite{james2013} among the coefficients in $\beta$ due to the limited number of perturbations. For instance, certain superpixels happen to be always masked or bunched together during the random perturbation, and thus their corresponding coefficients are correlated. By modeling $\lambda$, which works as a regularization term, we can mitigate the inflation on the standard error of the coefficient estimates. Consequently, EBLIME is more likely to assign correct positive and negative importance to the superpixels. On the other hand, $\lambda$ also adds flexibility to $\textrm{Cov}(\beta)$ and can lead to better uncertainty quantification. 

\subsubsection{Credible interval (CI) analysis}
As mentioned in \cref{prelim}, a major advantage of EBLIME and BayesLIME over LIME is that they allow us to infer the posterior distribution of $\beta$ with much less perturbations than computing the true value of $\beta$ with $2^p$ perturbations. Therefore, in BayesLIME \cite{bayeslime} the authors assessed the quality of the uncertainty estimates by checking how often the 95\% CIs derived from $P(\beta|Y)$ contain the true value of $\beta$. According to the definition, the 95\% CIs should include the ground truth 95\% of the time. Moreover, they claimed that the ground truth of $\beta$ can be approximated by the result of LIME under $10,000$ perturbed instances. This claim seems to be reasonable if the number of superpixels in the input image is less than $13$ (i.e., $p\leq13$). 

Here, we conduct the same procedure to compare BayesLIME and EBLIME, using all the benchmark datasets mentioned in BayesLIME. For illustration convenience, we elaborate on the result based on 100 ImageNet ``french bulldog'' images (results regarding the other datasets are included in \cref{other_dataset}). Specifically, each image is first segmented into around 13 superpixels via the default segmentation method in LIME \cite{lime}. Then, for each image, we obtain the true value of $\beta_j$ ($j=1,...,p$) using LIME and the 95\% CI for $\beta_j$ using BayesLIME or EBLIME. Based on the results of all the 100 images, we finally compute the fraction of the true $\beta_j$'s that fall within the 95\% CIs. We also tried different $N$ ranging from 50 to 500 to generate the 95\% CIs. This is to evaluate how size of the perturbation dataset $Z$ affects the quality of the posterior inference. In addition, this procedure is repeated five times but each time with a different random initialization to take account of the randomness in generating $Z$. 
% It is worth noting that we also conducted the same experiment on 100 images from another ImageNet class ``corn'', the result is similar and is included in \cref{appendix_B}.

\begin{figure}[hbt]
\begin{center}
\centerline{\includegraphics[width=\columnwidth]{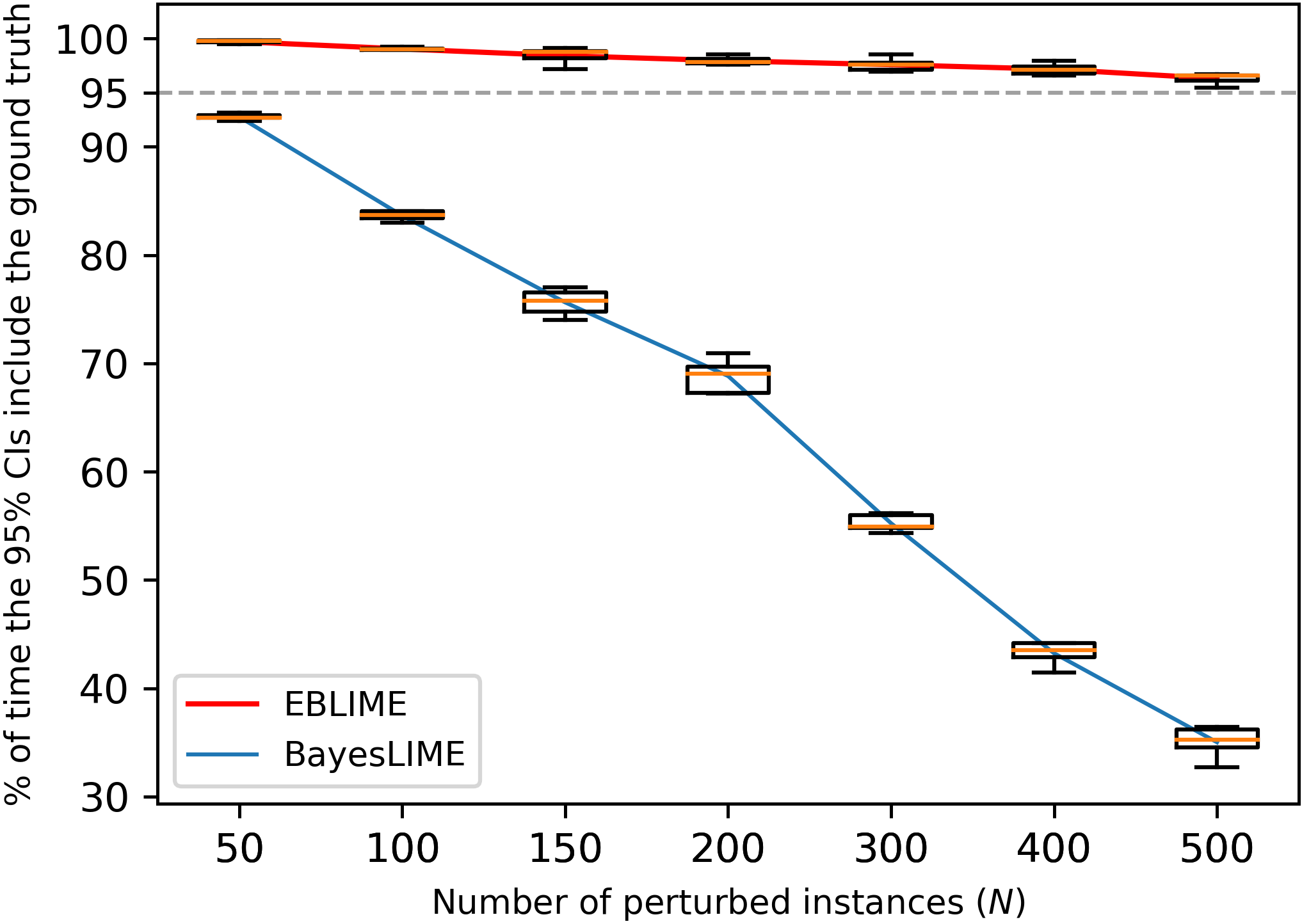}}
\caption{The percent of time the 95\% CIs (generated by EBLIME and BayesLIME under different $N$) include the ground truth of $\beta$ (generated by LIME with 10,000 perturbed instances). Closer to 95\% is better. At each $N$, the boxplot contains results from five different random initializations.}
\label{CIs}
\end{center}
% \vskip -0.1in
\end{figure}

The results are shown in \cref{CIs}. It can be observed that as the number of perturbed instances increases from 50 to 500, the 95\% CIs produced by BayesLIME include the ground truth less often (the percent of time drops from around 93\% to around 35\%). Similar patterns can also be found in \cref{other_CIs} in \cref{other_dataset}.

A potential reason is that the posterior mean estimated by BayesLIME deviated from the ground truth. When $N=50$, the 95\% CI is relatively wide so that it can cover the ground truth about 93\% of the time even with a deviated posterior mean. However, the 95\% CI would become narrower as $N$ increases because there is less uncertainty. Consequently, the 95\% CI from BayesLIME may converge to a wrong interval. On the other hand, EBLIME appears to be more conservative as the 95\% CIs always include the true value of $\beta$ slightly more than 95\% of the time. Moreover, the result is much more consistent than BayesLIME under different $N$ and random initializations.

\subsection{Localizing defects in 3D-printed components from ultrasonic images}
Currently, industrial-scale inspection of bulk products to locate internal defects such as pores and blowholes remains slow, expensive, and imprecise. Ultrasound methods are considered the most promising for bulk inspection of manufactured products \cite{brinksmeier1984nondestructive}. However, the sensitivity of ultrasonic inspection tends to be poor in many polymeric and composite materials \cite{bakaric2021}. 

To fill in this gap, multiple efforts are underway to combine the conventional physics-based image reconstruction with supervised machine learning methods, and thereby enhance the power of ultrasonic inspection \cite{gardner2020machine}. Nevertheless, they are limited to classifying an image for the presence or absence of a defect in a binary fashion, or segmenting the regions containing subsurface defects from the images. They can hardly localize or detect defects that are located at depths where the noise effects occlude the identity of the defects or create false-positive artifacts.

\begin{figure}[ht]
\begin{center}
\centerline{\includegraphics[width=\columnwidth]{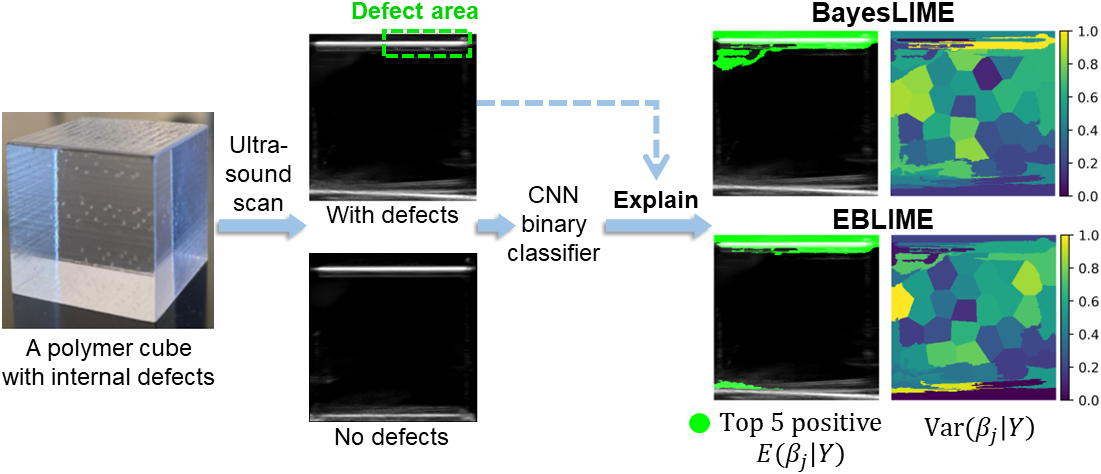}}
\caption{Workflow of explanation-infused ultrasonic internal defect localization with an example of BayesLIME and EBLIME results. The top 5 positively important superpixels (highlighted in green, corresponding to the five largest components by positive value in $E(\beta|Y)$) from both methods correctly localized the defect area. However, BayesLIME assigns a larger variance to these superpixels while EBLIME is more confident about the correct localization.}
\label{XAI_def_loc}
\end{center}
% \vskip -0.1in
\end{figure}

LIME's ability of identifying the most dominant superpixels in image classification can bring orders of magnitude increase in the speed and performance of industrial quality inspection methods for defect localization. Recently, Karthikeyan et~al. \yrcite{karthikeyan2022} proposed to use LIME to localize internal defect distributed in polymeric materials. These defects were introduced at specific locations using a 3D-printing technique. \cref{XAI_def_loc} illustrates the scheme of the explanation-infused ultrasonic internal defect localization. First, the polymer cube with internal artifacts undergoes an ultrasound scan to generate ultrasound images. Each ultrasound image can be labeled as ``With defects'' and ``No defects'' depending on whether that image is taken from a location where a defect was introduced in the 3D-printed part. As seen in \cref{XAI_def_loc}, the difference between the two classes can be hard to discern from the ultrasound image with the naked eye. Hence automated models such as CNN come in handy for future inspection. By explaining a ``With defects'' ultrasound image, we can identify its most positively important superpixels towards the prediction of ``With defects''. If the CNN model has indeed learned the defect patterns, these superpixels are supposed to localize the defect and other areas affected by the existence of the defect. 

Based on the workflow, we evaluate the defect localization performance of EBLIME and compare it with LIME and BayesLIME. First, a five-layer CNN binary classifier was built upon 123 ultrasound images with both 100\% training and test accuracy. Then, LIME, BayesLIME, and EBLIME were applied to explain the model's predictions on 69 ``With defect'' images. During the implementation, each ultrasound image was segmented into around 40 superpixels and other settings were kept the same as in the first case study. \cref{XAI_def_loc} shows an example of the defect localization result where EBLIME provides a better uncertainty quantification than BayesLIME. 

Overall, for each method, we count the number of images where the defect area is correctly covered by the top 5 positive superpixels. The result (see \cref{table1}) suggests that EBLIME correctly localized defects in 2.9\% more images than LIME and BayesLIME. The performance can be further improved with more ultrasound images of the product. It is also noteworthy that the ground truth in terms of the locations of the defects, i.e., the most important superpixels, is known in this case study. The practical industrial importance together with this information make this case study a candidate benchmark for future investigations into post-hoc explanation methods. 

\begin{table}[hbt]
\caption{Internal defect localization results.}
\label{table1}
\vskip 0.1in
\begin{center}
\begin{small}
\begin{sc}
\begin{tabular}{lcccr}
\toprule
Method & Correct localizations \\ 
\midrule
LIME            & 48/69 (69.6\%)                 \\
BayesLIME       & 48/69 (69.6\%)                 \\
EBLIME          & \textbf{50/69 (72.5\%)}                 \\ 
\bottomrule
\end{tabular}
\end{sc}
\end{small}
\end{center}
\vskip -0.1in
\end{table}

\subsection{Significance of \texorpdfstring{$\lambda$}{lambda}}
\begin{figure}[ht]
\vskip 0.1in
\begin{center}
\centerline{\includegraphics[width=0.66\columnwidth]{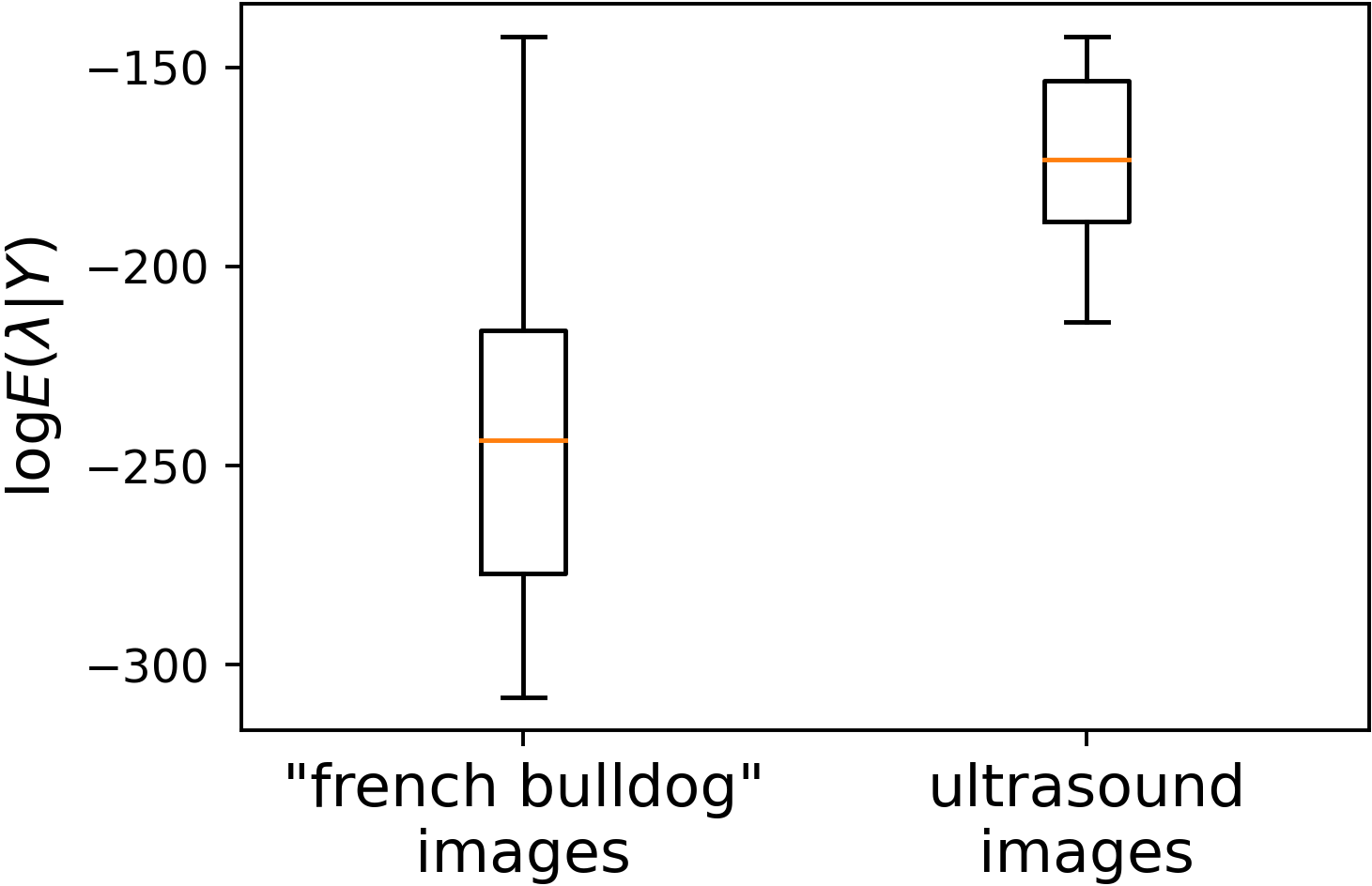}}
\caption{$\log{E(\lambda|Y)}$ of the good (intuitive and correct) explanation results from the two case studies.}
\label{lam_boxplot}
\end{center}
% \vskip -0.1in
\end{figure}

We computed the posterior mean of $\lambda$ based on \cref{prop2} for selected good explanation results from both case studies. In particular, in the first case study, we selected 100 images where the top superpixels correctly include dog features as positively important and others as negatively important. In the second case, we picked the 32 images where the defect is correctly localized by EBLIME. As evident in \cref{lam_boxplot}, $E(\lambda|Y)$ varies when explaining different input images. This aligns with our \cref{r1} that the ridge parameter $\lambda$ is not supposed to be set as a constant.

\section{Conclusion}\label{conclusion}
We proposed a Bayesian ridge regression approach---EBLIME to explain black-box machine learning models, i.e., quantify features' contribution (importance) to the model's prediction on a given input as well as the underlying uncertainty. The method aims to enhance a prior Bayesian explanation method in the following way. The prior method assumes that the feature importance $\beta$ and the error of the linear explanation model $\epsilon$ share the same uncertainty. This assumption has been shown to severely impede the performance of the earlier method in certain cases. In contrast, EBLIME had aimed to improve the explanation performance by introducing a random variable, a regularization parameter $\lambda$ to scale the covariance of $\beta$. 

To evaluate the performance of EBLIME and compare it with the state-of-the-art methods, two case studies were conducted, namely, on a benchmark dataset as well as a real-world application of internal defect localization.  Overall, our results indicate that EBLIME can outperform the existing methods, especially in terms of generating the posterior distribution, credible intervals, and feature rankings. Consequently, it can more effectively distinguish positively important superpixels (e.g., the major object in the image) from the negatively important ones (e.g., background), especially in cases where the prior methods may fail. This can be a potential advantage for object detection applications.

The detailed evaluation of EBLIME on tabular and textual data can be included in future study. Further efforts can also be made to develop ranking criteria for Bayesian explanation methods to compare the feature importance distributions. 

\bibliography{example_paper}

\begin{thebibliography}{25}
\providecommand{\natexlab}[1]{#1}
\providecommand{\url}[1]{\texttt{#1}}
\expandafter\ifx\csname urlstyle\endcsname\relax
  \providecommand{\doi}[1]{doi: #1}\else
  \providecommand{\doi}{doi: \begingroup \urlstyle{rm}\Url}\fi

\bibitem[{Acumen Research and Consulting}(2022)]{acumen2022}
{Acumen Research and Consulting}.
\newblock Deep learning market size will achieve usd 415 billion by 2030
  growing at 51.1\% cagr fueled by the increasing adoption of big data
  analytics, Aug 2022.
\newblock URL
  \url{https://www.globenewswire.com/news-release/2022/08/16/2499313/0/en/Deep-Learning-Market-Size-Will-Achieve-USD-415-Billion-by-2030-growing-at-51-1-CAGR-Fueled-by-the-Increasing-Adoption-of-Big-Data-Analytics-Exclusive-Report-by-Acumen-Research-and-.html}.

\bibitem[Bakaric et~al.(2021)Bakaric, Miloro, Javaherian, Cox, Treeby, and
  Brown]{bakaric2021}
Bakaric, M., Miloro, P., Javaherian, A., Cox, B.~T., Treeby, B.~E., and Brown,
  M.~D.
\newblock Measurement of the ultrasound attenuation and dispersion in
  3{D}-printed photopolymer materials from 1 to 3.5 mhz.
\newblock \emph{The Journal of the Acoustical Society of America}, 150\penalty0
  (4):\penalty0 2798--2805, 2021.

\bibitem[Bishop \& Nasrabadi(2006)Bishop and Nasrabadi]{bishop2006}
Bishop, C.~M. and Nasrabadi, N.~M.
\newblock \emph{Pattern recognition and machine learning}, volume~4.
\newblock Springer, 2006.

\bibitem[Brinksmeier et~al.(1984)Brinksmeier, Schneider, Theiner, and
  T{\"o}nshoff]{brinksmeier1984nondestructive}
Brinksmeier, E., Schneider, E., Theiner, W., and T{\"o}nshoff, H.
\newblock Nondestructive testing for evaluating surface integrity.
\newblock \emph{CIRP Annals}, 33\penalty0 (2):\penalty0 489--509, 1984.

\bibitem[Deng et~al.(2009)Deng, Dong, Socher, Li, Li, and Fei-Fei]{deng2009}
Deng, J., Dong, W., Socher, R., Li, L.-J., Li, K., and Fei-Fei, L.
\newblock Image{N}et: A large-scale hierarchical image database.
\newblock In \emph{2009 IEEE Conference on Computer Vision and Pattern
  Recognition}, pp.\  248--255. IEEE, 2009.

\bibitem[Gardner et~al.(2020)Gardner, Fuentes, Dervilis, Mineo, Pierce, Cross,
  and Worden]{gardner2020machine}
Gardner, P., Fuentes, R., Dervilis, N., Mineo, C., Pierce, S., Cross, E., and
  Worden, K.
\newblock Machine learning at the interface of structural health monitoring and
  non-destructive evaluation.
\newblock \emph{Philosophical Transactions of the Royal Society A},
  378\penalty0 (2182):\penalty0 20190581, 2020.

\bibitem[Gelman(2006)]{sigmaprior}
Gelman, A.
\newblock Prior distributions for variance parameters in hierarchical models
  (comment on article by {B}rowne and {D}raper).
\newblock \emph{Bayesian Analysis}, 1\penalty0 (3):\penalty0 515--534, 2006.

\bibitem[Gramegna \& Giudici(2021)Gramegna and Giudici]{gramegna2021}
Gramegna, A. and Giudici, P.
\newblock {SHAP} and {LIME}: An evaluation of discriminative power in credit
  risk.
\newblock \emph{Frontiers in Artificial Intelligence}, pp.\  140, 2021.

\bibitem[Guo et~al.(2018)Guo, Huang, Tao, Xing, and Lin]{guo2018}
Guo, W., Huang, S., Tao, Y., Xing, X., and Lin, L.
\newblock Explaining deep learning models--a {B}ayesian non-parametric
  approach.
\newblock \emph{Advances in Neural Information Processing Systems}, 31, 2018.

\bibitem[Higham(2002)]{higham2002}
Higham, N.~J.
\newblock \emph{Accuracy and stability of numerical algorithms}.
\newblock SIAM, 2002.

\bibitem[James et~al.(2013)James, Witten, Hastie, and Tibshirani]{james2013}
James, G., Witten, D., Hastie, T., and Tibshirani, R.
\newblock \emph{An introduction to statistical learning}, volume 112.
\newblock Springer, 2013.

\bibitem[Johndrow et~al.(2020)Johndrow, Orenstein, and
  Bhattacharya]{johndrow2020}
Johndrow, J., Orenstein, P., and Bhattacharya, A.
\newblock Scalable approximate {MCMC} algorithms for the horseshoe prior.
\newblock \emph{Journal of Machine Learning Research}, 21\penalty0 (73), 2020.

\bibitem[Karthikeyan et~al.(2022)Karthikeyan, Tiwari, Zhong, and
  Bukkapatnam]{karthikeyan2022}
Karthikeyan, A., Tiwari, A., Zhong, Y., and Bukkapatnam, S.~T.
\newblock Explainable {AI}-infused ultrasonic inspection for internal defect
  detection.
\newblock \emph{CIRP Annals}, 2022.

\bibitem[Maddison et~al.(2016)Maddison, Mnih, and Teh]{maddison2016}
Maddison, C.~J., Mnih, A., and Teh, Y.~W.
\newblock The concrete distribution: A continuous relaxation of discrete random
  variables.
\newblock \emph{arXiv preprint arXiv:1611.00712}, 2016.

\bibitem[Polson \& Scott(2012)Polson and Scott]{polson2012}
Polson, N.~G. and Scott, J.~G.
\newblock On the half-{C}auchy prior for a global scale parameter.
\newblock \emph{Bayesian Analysis}, 7\penalty0 (4):\penalty0 887--902, 2012.

\bibitem[Ribeiro et~al.(2016)Ribeiro, Singh, and Guestrin]{lime}
Ribeiro, M.~T., Singh, S., and Guestrin, C.
\newblock "{W}hy should i trust you?" {E}xplaining the predictions of any
  classifier.
\newblock In \emph{Proceedings of the 22nd ACM SIGKDD International Conference
  on Knowledge Discovery and Data Mining}, pp.\  1135--1144, 2016.

\bibitem[Saini \& Prasad(2022)Saini and Prasad]{saini2022}
Saini, A. and Prasad, R.
\newblock Select wisely and explain: Active learning and probabilistic local
  post-hoc explainability.
\newblock In \emph{Proceedings of the 2022 AAAI/ACM Conference on AI, Ethics,
  and Society}, pp.\  599--608, 2022.

\bibitem[Simonyan \& Zisserman(2014)Simonyan and Zisserman]{simonyan2014}
Simonyan, K. and Zisserman, A.
\newblock Very deep convolutional networks for large-scale image recognition.
\newblock \emph{arXiv preprint arXiv:1409.1556}, 2014.

\bibitem[Slack et~al.(2021)Slack, Hilgard, Singh, and Lakkaraju]{bayeslime}
Slack, D., Hilgard, A., Singh, S., and Lakkaraju, H.
\newblock Reliable post hoc explanations: Modeling uncertainty in
  explainability.
\newblock \emph{Advances in Neural Information Processing Systems},
  34:\penalty0 9391--9404, 2021.

\bibitem[Visani et~al.(2022)Visani, Bagli, Chesani, Poluzzi, and
  Capuzzo]{visani2022}
Visani, G., Bagli, E., Chesani, F., Poluzzi, A., and Capuzzo, D.
\newblock Statistical stability indices for {LIME}: Obtaining reliable
  explanations for machine learning models.
\newblock \emph{Journal of the Operational Research Society}, 73\penalty0
  (1):\penalty0 91--101, 2022.

\bibitem[Zafar \& Khan(2019)Zafar and Khan]{zafar2019dlime}
Zafar, M.~R. and Khan, N.~M.
\newblock Dlime: A deterministic local interpretable model-agnostic
  explanations approach for computer-aided diagnosis systems.
\newblock \emph{arXiv preprint arXiv:1906.10263}, 2019.

\bibitem[Zhang et~al.(2019)Zhang, Song, Sun, Tan, and Udell]{zhang2019}
Zhang, Y., Song, K., Sun, Y., Tan, S., and Udell, M.
\newblock "{W}hy should you trust my explanation?" {U}nderstanding uncertainty
  in {LIME} explanations.
\newblock \emph{arXiv preprint arXiv:1904.12991}, 2019.

\bibitem[Zhao et~al.(2021)Zhao, Huang, Huang, Robu, and Flynn]{zhao2021}
Zhao, X., Huang, W., Huang, X., Robu, V., and Flynn, D.
\newblock Baylime: Bayesian local interpretable model-agnostic explanations.
\newblock In \emph{Uncertainty in Artificial Intelligence}, pp.\  887--896.
  PMLR, 2021.

\bibitem[Zhong et~al.(2022)Zhong, Tiwari, Yamaguchi, Lakhtakia, and
  Bukkapatnam]{zhong2022}
Zhong, Y., Tiwari, A., Yamaguchi, H., Lakhtakia, A., and Bukkapatnam, S.~T.
\newblock Identifying the influence of surface texture waveforms on colors of
  polished surfaces using an explainable {AI} approach.
\newblock \emph{IISE Transactions}, pp.\  1--15, 2022.

\bibitem[Zhou et~al.(2021)Zhou, Hooker, and Wang]{zhou2021}
Zhou, Z., Hooker, G., and Wang, F.
\newblock S-lime: Stabilized-{LIME} for model explanation.
\newblock In \emph{Proceedings of the 27th ACM SIGKDD Conference on Knowledge
  Discovery \& Data Mining}, pp.\  2429--2438, 2021.

\end{thebibliography}
\bibliographystyle{icml2023}

%%%%%%%%%%%%%%%%%%%%%%%%%%%%%%%%%%%%%%%%%%%%%%%%%%%%%%%%%%%%%%%%%%%%%%%%%%%%%%%
%%%%%%%%%%%%%%%%%%%%%%%%%%%%%%%%%%%%%%%%%%%%%%%%%%%%%%%%%%%%%%%%%%%%%%%%%%%%%%%
% APPENDIX
%%%%%%%%%%%%%%%%%%%%%%%%%%%%%%%%%%%%%%%%%%%%%%%%%%%%%%%%%%%%%%%%%%%%%%%%%%%%%%%
%%%%%%%%%%%%%%%%%%%%%%%%%%%%%%%%%%%%%%%%%%%%%%%%%%%%%%%%%%%%%%%%%%%%%%%%%%%%%%%
\newpage
\appendix
\onecolumn
\section*{Appendix} 
\section{Proof of propositions}\label{proofs}
\textbf{\cref{prop0}}\hspace{3pt} Using Bayesian conjugacy adapted to the weighted ridge regression setup, we can write
\begin{gather*}
    \beta|\sigma^2, \lambda, Y \sim \mathcal{N}(\hat{\beta}, V_\lambda \sigma^2) \\
    \sigma^2|\lambda,Y \sim \textit{\textrm{Inverse-Gamma}}(a+N/2,Q_{\lambda}/2) 
\end{gather*}
wherein $Q_{\lambda} = Y^T\big(\normalfont{\textrm{diag}}^{-1}(\Pi_x(Z))+\lambda^{-1}ZZ^T\big)^{-1}Y + 2b$. 
\begin{proof}
According to normal distribution theory, \cref{eq2sub1}, \cref{eq2sub2}, and \cref{eq2sub3}, we get $Y|\sigma^2,\lambda \sim \mathcal{N}(0, M_\lambda\sigma^2)$, wherein $M_\lambda=\normalfont{\textrm{diag}}^{-1}(\Pi_x(Z))+\lambda^{-1}ZZ^T$. Next,
\begin{equation*}
\begin{aligned}
P(\beta|\sigma^2, \lambda, Y)
    &=  \frac{P(Y|\beta,\sigma^2,\lambda)P(\beta|\sigma^2,\lambda)}{P(Y|\sigma^2, \lambda)} \\
    &\propto \exp{\Big[(-\frac{1}{2\sigma^2})\big((Y-Z\beta)^T \textrm{diag}(\Pi_x(Z))(Y-Z\beta)+\beta^T(\lambda^{-1}\mathbb{I}_p)^{-1}\beta-(Y^T{M_\lambda}^{-1}Y)\big)\Big]} \\
    &\propto \exp{[(-\frac{1}{2\sigma^2})\Big(\beta^T(Z^T\textrm{diag}(\Pi_x(Z))Z+\lambda\mathbb{I}_p)\beta - 2Y^T\textrm{diag}(\Pi_x(Z))Z\beta - Y^TM_\lambda^{-1})Y\Big)]}
\end{aligned}
\end{equation*}
By plugging in $\hat{\beta}$, completing the square $(\beta-\hat{\beta})^2$ and using Woodbury formula \cite{higham2002} on ${M_\lambda}^{-1}$, we can derive \cref{beta_cond_pos}. 

Since $P(\sigma^2|\lambda,Y) \propto P(Y|\sigma^2,\lambda) P(\sigma^2) \propto (\sigma^2)^{-(a+N/2)-1}\exp[-(Y^TM_{\lambda}^{-1}Y+2b)/(2\sigma^2)]$, we can derive \cref{sigma_cond_pos}.
\end{proof}

\textbf{\cref{prop2}}\hspace{3pt} The posterior density of $\lambda$ can be written as
\begin{flalign*}  
    \begin{aligned} 
    P(\lambda|Y)
    & \propto |M_{\lambda}|^{-1/2}(Q_{\lambda})^{-(a+N/2)}P(\lambda).
    \end{aligned} 
\end{flalign*}
Its posterior mean can be approximated by discretizing $P(\lambda|Y)$ over $\{\lambda_1,...,\lambda_L\}$, that is
\begin{equation*} 
    {E(\lambda|Y)} \approx \sum^{L}_{l=1}\lambda_l|M_{\lambda_l}|^{-1/2}(Q_{\lambda_l})^{-(a+N/2)}P'(\lambda_l)
\end{equation*} 
 % $|\cdot|$ denotes the matrix determinant
wherein $L$ is the number of discretized $\lambda$ values and $P'(\lambda)$ denotes the normalized $P(\lambda)$ which satisfies $\sum^{L}_{l=1}P'(\lambda_l)=1$.

\begin{proof} $P(\lambda|Y) \propto P(Y|\lambda)P(\lambda) = \int P(Y|\sigma^2,\lambda)P(\lambda)P(\sigma^2)d\sigma^2$. Here, the integration over $\sigma^2$ can be computed analytically by virtue of the inverse-gamma conjugate prior and change of variables. Specifically,
\begin{equation}
\begin{aligned}
    P(\lambda|Y) \propto P(Y|\lambda)P(\lambda) 
    &= \int P(Y|\sigma^2,\lambda)P(\lambda)P(\sigma^2)d\sigma^2\\
    &\propto \int |M_\lambda\sigma^2|^{-\frac{1}{2}}\exp{(-\frac{Y^T{M_\lambda}^{-1}Y}{2\sigma^2})}(\sigma^2)^{-a-1}\exp{(-\frac{b}{\sigma^2})}P(\lambda)d\sigma^2 \\
    &= \int |M_\lambda|^{-\frac{1}{2}}P(\lambda)(\sigma^2)^{-a-1-\frac{N}{2}}\exp{(-\frac{Q_\lambda}{2\sigma^2})} d\sigma^2
\end{aligned}
\end{equation}
Let $\theta =Q_\lambda/(2\sigma^2)$, we have $\sigma^2 = Q_\lambda/(2\theta)$ and $d\sigma^2/d\theta = -Q_\lambda/(2\theta^2)$. Now we write the above integral as
\begin{equation*}
\begin{aligned}
    &= |M_\lambda|^{-\frac{1}{2}}P(\lambda) \int\frac{\theta^{a+1+\frac{N}{2}}}{(\frac{Q_\lambda}{2})^{a+1+\frac{N}{2}}} [-\frac{1}{\theta^2}(\frac{Q_\lambda}{2})]\exp{(-\theta)}d\theta \\
    &\propto |M_\lambda|^{-\frac{1}{2}}P(\lambda)(Q_\lambda)^{-(a+\frac{N}{2})}\int\theta^{a-1+\frac{N}{2}}\exp{(-\theta)}d\theta \\
    &\propto |M_\lambda|^{-\frac{1}{2}}(Q_\lambda)^{-(a+\frac{N}{2})}P(\lambda)
\end{aligned}
\end{equation*}

The posterior mean $E(\lambda|Y) = \int \lambda P(\lambda|Y)d\lambda \propto \int \lambda |M_{\lambda}|^{-1/2}(Q_{\lambda})^{-(a+N/2)}P(\lambda)d\lambda$. Here, the integration over $\lambda$ is approximated by discretization. In particular, we sample $\{\lambda_1,...,\lambda_L\}$ uniformly over the range $(0,r)$. $E(\lambda|Y)$ can then be obtained as \cref{lam_pos_mean}.
\end{proof}

\section{Credible interval analysis on other benchmark datasets}\label{other_dataset}
We provide the credible interval analysis results on all the datasets used in BayesLIME. This includes tabular datasets “compass” and “german credit”, image sets “corn”, “broccoli”, “scuba diver” from ImageNet, and image sets “Digit 1-9” from MNIST. 

We can again observe that BayesLIME 95\% CIs include the ground truth much less often as $N$ increases, except for two datasets: “Digit 9” image set and “German credit” tabular dataset. On the other hand, EBLIME is more conservative and consistent (i.e., no rapid drop) for all datasets. 

\begin{table}[hbt]
\caption{\% of time the 95\% CIs (generated by EBLIME and BayesLIME under different $N$) include the ground truth of $\beta$. Closer to 95\% is better.}
\label{other_CIs}
\begin{center}
\begin{tabular}{c|ccc|ccc}
\hline
 & \multicolumn{3}{c|}{\textbf{BayesLIME}}                & \multicolumn{3}{c}{\textbf{EBLIME}}                    \\
 & \textbf{$N$ = 100} & \textbf{$N$ = 200} & \textbf{$N$ = 400} & \textbf{$N$ = 100} & \textbf{$N$ = 200} & \textbf{$N$ = 400} \\ \hline
\textbf{Image dataset}   &        &        &        &         &         &         \\
French bulldog           & 84.1\% & 69.7\% & 43.5\% & 98.9\%  & 98.1\%  & 97.4\%  \\
Broccoli                 & 81.0\% & 66.1\% & 47.1\% & 100.0\% & 100.0\% & 100.0\% \\
Corn                     & 85.1\% & 71.5\% & 50.1\% & 100.0\% & 99.4\%  & 98.7\%  \\
Scuba diver              & 87.5\% & 73.2\% & 61.2\% & 99.4\%  & 97.2\%  & 95.6\%  \\
Digit 1                  & 89.9\% & 81.8\% & 64.2\% & 99.2\%  & 97.8\%  & 96.6\%  \\
Digit 2                  & 93.3\% & 90.0\% & 83.8\% & 98.7\%  & 97.5\%  & 95.8\%  \\
Digit 3                  & 90.7\% & 83.9\% & 68.5\% & 98.6\%  & 97.9\%  & 96.7\%  \\
Digit 4                  & 93.2\% & 92.3\% & 87.9\% & 99.0\%  & 97.6\%  & 96.2\%  \\
Digit 5                  & 91.1\% & 82.3\% & 70.5\% & 98.8\%  & 97.2\%  & 96.8\%  \\
Digit 6                  & 93.8\% & 93.3\% & 90.9\% & 97.8\%  & 97.0\%  & 95.8\%  \\
Digit 7                  & 90.6\% & 85.8\% & 76.8\% & 98.8\%  & 97.5\%  & 96.4\%  \\
Digit 8                  & 93.4\% & 93.2\% & 90.2\% & 97.4\%  & 96.8\%  & 95.1\%  \\
Digit 9                  & 94.5\% & 92.9\% & 92.8\% & 98.8\%  & 96.4\%  & 96.8\%  \\
\textbf{Tabular dataset} &        &        &        &         &         &         \\
Compas                   & 45.7\% & 22.7\% & 9.8\%  & 96.9\%  & 95.8\%  & 95.7\%  \\
German credit            & 90.5\% & 90.6\% & 91.5\% & 99.8\%  & 99.4\%  & 98.1\%  \\ \hline
\end{tabular}
\end{center}
\end{table}

%%%%%%%%%%%%%%%%%%%%%%%%%%%%%%%%%%%%%%%%%%%%%%%%%%%%%%%%%%%%%%%%%%%%%%%%%%%%%%%
%%%%%%%%%%%%%%%%%%%%%%%%%%%%%%%%%%%%%%%%%%%%%%%%%%%%%%%%%%%%%%%%%%%%%%%%%%%%%%%

\end{document}